\pdfoutput=1

\documentclass[11pt]{article}

\usepackage[]{ACL2023}

\usepackage{times}
\usepackage{latexsym}
\usepackage{graphicx}
\usepackage[T1]{fontenc}
\usepackage{amsfonts,amssymb}
\usepackage{amsmath}
\usepackage{mathrsfs}
\usepackage{cleveref}

\usepackage[utf8]{inputenc}

\usepackage{microtype}

\usepackage{inconsolata}
\usepackage{multirow}
\usepackage{arydshln}
\usepackage{booktabs} 
\usepackage{pifont}

\usepackage[T1]{fontenc}

\usepackage[utf8]{inputenc}

\usepackage{microtype}

\usepackage{inconsolata}

\title{Unified Language Representation for Question Answering \\ over Text, Tables, and Images
}
\author{
  Bowen Yu,
  Cheng Fu,
  Haiyang Yu,
  Fei Huang,
  Yongbin Li\textsuperscript{\thanks{\quad Corresponding author.}}\\
  DAMO Academy, Alibaba Group \\
  {\tt \{yubowen.ybw,fucheng.fuc,yifei.yhy,f.huang,shuide.lyb\}@alibaba-inc.com} 
}

\begin{document}
\maketitle
\begin{abstract}

When trying to answer complex questions, people often rely on multiple sources of information, such as visual, textual, and tabular data. 
Previous approaches to this problem have focused on designing input features or model structure in the multi-modal space, which is inflexible for cross-modal reasoning or data-efficient training. 
In this paper, we call for an alternative paradigm, which transforms the images and tables into unified language representations, so that we can simplify the task into a simpler textual QA problem that can be solved using three steps: retrieval, ranking, and generation, all within a language space. 
This idea takes advantage of the power of pre-trained language models and is implemented in a framework called Solar. 
 Our experimental results show that Solar outperforms all existing methods by 10.6-32.3 pts on two datasets, MultimodalQA and MMCoQA, across ten different metrics. Additionally, Solar achieves the best performance on the WebQA leaderboard\footnote{The code of this work is available at \url{https://github.com/AlibabaResearch/DAMO-ConvAI/tree/main/solar}.}.

\end{abstract}

\section{Introduction}


Information overload is a major problem in today's society due to the vast amount of information available. 
Question answering (QA) systems can help alleviate this problem by providing users with concise and accurate answers\footnote{In this paper, we present a uniform solution to  question answering and conventional question answering. To avoid confusion, we will refer to these two tasks as QA.}~\cite{rajpurkar2016squad}. 
However, traditional QA models are limited to text or structured data and are unable to utilize the vast amount of multimodal knowledge available on the internet~\cite{hannan2020manymodalqa}, much of which is in non-text formats, such as images~\cite{talmor2020multimodalqa}, data tables~\cite{zhu2021tat}, etc.
For example, on the Wikipedia page for the United States, a user's questions about the US census, the color of the Statue of Liberty, and the US capital city could be answered using data tables, images, and paragraphs on the page, respectively.


To address this challenge, one strategy is to train separate QA models for each modality and use a classifier to determine which modality to obtain information from~\cite{talmor2020multimodalqa, li2022mmcoqa}.
While this method is straightforward, it lacks the ability of cross-modal reasoning as models of different modalities are integrated without interaction.
However, both human thought and the current benchmarks require reasoning across modalities~\cite{talmor2020multimodalqa}.
An alternative approach is to train a multi-modal model that can accept input from multiple sources simultaneously and produce the answer directly~\cite{chen2022murag, yang2022enhancing}. 
This method allows for reasoning between modalities, but it can be challenging to train due to the diversified inputs (e.g. text, image, and table can be combined in seven modal combinations). 
For example, MuRAG~\cite{chen2022murag} requires 3 million image-text pairs for pre-training to achieve decent results.

\begin{figure*}
    \centering
    \includegraphics[width=0.95\linewidth]{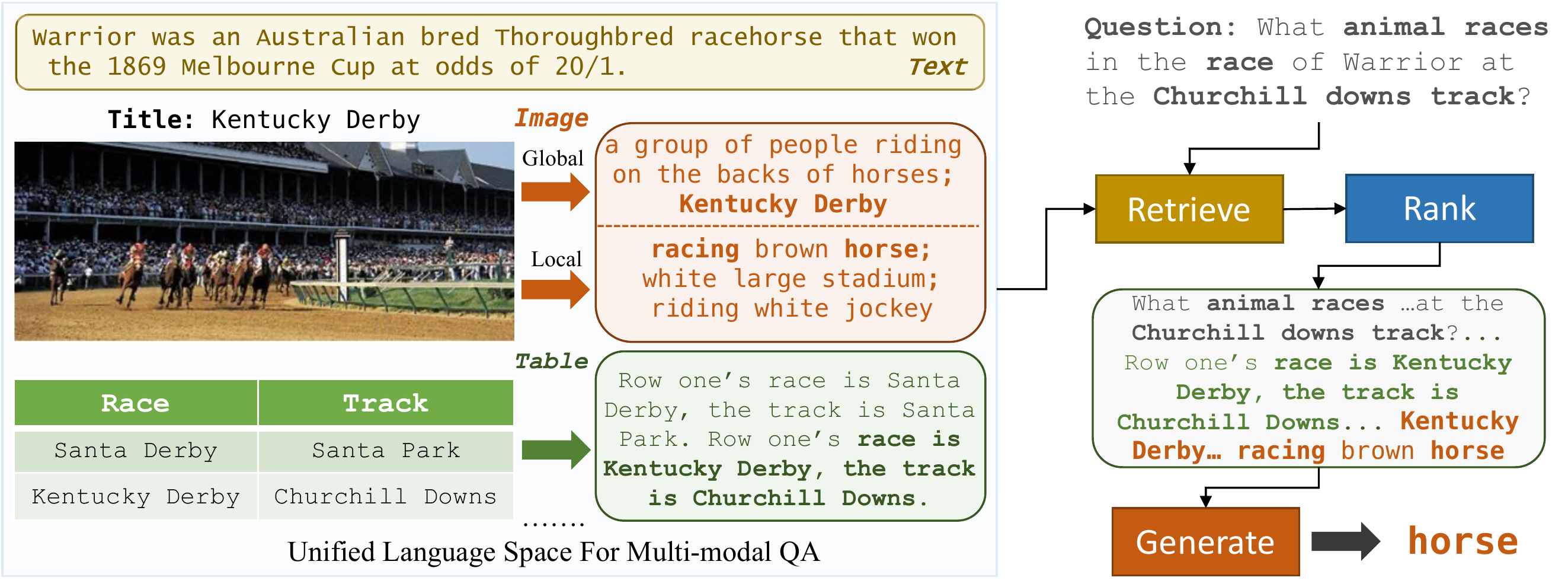}
    \caption{A schematic overview of our approach.}
    \label{fig:model}
    \vspace{-0.1in}
\end{figure*}

In this paper, we introduce Solar, a unified language representation for multi-modal QA. 
Unlike existing QA methods performing in the multi-modal space, Solar transforms images and tables into natural languages at the input stage, simplifying the multi-modal QA task into a simpler textual QA problem. 
This transformation has at least three advantages as follows:
(1) it helps to overcome the diversified modal combinations and enable cross-modal reasoning by putting together the textual representation of relevant clues belonging to multiple modalities;
(2) language models have learned an enormous amount of world knowledge and patterns by reading billion-level text corpora~\cite{hao2022language}, compared with the much smaller multi-modal pre-training datasets used in MuRAG.
The memorized information can serve as reusable background knowledge and basic skills for complex QA~\cite{dai2022knowledge};
(3) Language has a higher information density compared to images~\cite{he2022masked} and tables, the transformation from low-density to high-density can be done losslessly for tables and with minimal loss for images using global and local textualization strategies. 
Then we retrieve relevant clues in the unified language space for each question, rank clues to get the most related top-$N$ ones, and input them into the decoding model to generate the desired answer.



Though simple, our experiments show that Solar achieves state-of-the-art performance on three different multi-modal QA datasets.
On MultiModalQA and MMCoQA, we outperform sophisticated baselines by 10.6-32.3 pts across ten metrics.
Without the need of pre-training on external vision-language data or additional tricks, Solar also achieves the first position on the official WebQA leaderboard at the time of writing\footnote{\url{https://eval.ai/web/challenges/challenge-page/1255/leaderboard/3168}}. 
At the same time, Solar has reduced 99.7\% of storage space.

\section{Method}
\label{sec:method}

In this section, we will introduce Solar in detail. As shown in Figure~\ref{fig:model}, it consists of two parts: a unified language representation and a unified QA model. 
The former is responsible for mapping images, tables and texts in different modal spaces into a unified language space. 
The latter generates an answer through three steps of retrieval, ranking and decoding in the language space. We discuss the related work in \textbf{Appendix}~\ref{sec:related}.

\subsection{Unified Language Representation}
\label{sec:method_representation}
Multi-modal QA often contains heterogeneous information sources of text, table and image. 
Therefore, if we want to implement QA in a unified language space, we need to first convert tables and images into text representations in the input layer.

\textbf{Table.} We use simple natural language templates to transform tables into sentences that sound natural to humans. 
As an example, consider the table in Figure~\ref{fig:model}. We can turn the table into a sentence by arranging the cells in a linear fashion, like this: "Row one's race is Santa Derby, the track is Santa Park. Row two is..." and so on. 
This process doesn't lose any information - the original structure of the table can be reconstructed from the resulting natural language sequence~\cite{chen2019tabfact}.


\textbf{Image.} Compared with tables, image has no natural language structure, so transforming image into natural language must be information damaging; In order to alleviate information loss, we propose two transformation strategies: global and local.
The global strategy involves using a trained image caption model to generate text descriptions of the macro scenes of the image~\cite{changpinyo2022all, yang2022empirical}, including the image's title if it has one. 
The local strategy leverages a trained object-attribute detection model~\cite{gao2022transform} to describe the individual objects and their attributes in the image such as "racing brown horse" in Figure~\ref{fig:model}, to compensate for missing detail semantics in the overall description.
Then the global and local sequences are spliced through punctations as the language representation of the image.



\textbf{Question.} We hope that Solar will be able to handle both standard QA and more complex conventional QA tasks that involve taking into account the conversation history. To accomplish this, we concatenate the current question with the previous questions and answers in the conversation to create a contextual question $q$ for the system to process.



\subsection{Unified Multi-Modal QA Model}
\label{sec:method_model}



After building the unified space $\mathbb{Z}$ containing the language representation of all heterogeneous sources, we can convert the multi-modal QA task into a simple textual QA problem.
To find the answer to a question $q$ from $\mathbb{Z}$, Solar follows these steps: first, it retrieves the $K$ clues in $\mathbb{Z}$ most relevant to the question; next, it accurately sorts these clues for the question to get the top-$N$ valuable clues; finally, it sends these clues to the decoding component to generate the answer.
The training details of these  steps are presented in \textbf{Appendix~\ref{sec:training_details}}


\textbf{Retrieve.} The first step is to retrieve several question-relevant clues in the language space $\mathbb{Z}$ to ground the following generation. 
We use DPR~\cite{karpukhin2020dense} as our retriever, which projects questions and textual clues to a shared space using BERT~\cite{kenton2019bert}). 
Formally, we retrieve $K$ most relevant clues $\mathbb{Z}_K=z_{[1, \cdots, K]}\in \mathbb{Z}$ for question $q$ as:
\begin{equation}
\small
        \mathbb{Z}_K = \left\{z_i\in \mathbb{Z}|\mathrm{topK}\, \{\textrm{BERT}(q)^\top \textrm{BERT}(z_i)\} \right\}.
\end{equation}
\textbf{Rank.} The ranker we use is based on the sequence-pair classification.
The question $q$ and each candidate clue $z_i \in \mathbb{Z}_K$ are input together to a BERT followed by a projection layer and Sigmoid function to calculate the ranking score of $z_i$:
\begin{equation}
    s_i={\rm Sigmoid}({\rm Linear}({\rm BERT} (z_i \oplus q))\}.
\end{equation}

Cross attention is applied over the tokens of both sequences jointly.
Therefore, compared with the independent coding of DPR, it can more accurately model the correlation between $q$ and $z_i$. 
In the overall architecture, the ranker plays two roles: first, to find the most relevant clue, which is one of the evaluation metrics; Then, delete irrelevant clues of error recall in the DPR stage to reduce noise in the input of decoding component.



\textbf{Generate.} The top-$N$ clues with the highest ranking scores in $\mathbb{Z}_K$ are then joined with $q$ and passed to a encoder-decoder network such as T5~\cite{raffel2020exploring} to generate the answer $a$: 
Note that the top-$N$ clues may contain related messages belonging to different modalities. Through the cross attention in the T5 transformer decoding, reasoning between and within modalities can be naturally realized.
For instance, consider the text sequence shown in the lower right corner of Figure~\ref{fig:model}. This sequence includes language representations of tables (highlighted in green) and images (highlighted in red), which are combined with the question to create a reasoning chain of \textit{Churchill downs track} -> \textit{Kentucky Derby} -> \textit{racing} -> \textit{horse}.

\subsection{Model Training}
\label{sec:training_details}

Our training is carried out in three phases: retrieval training, ranking training, and generation training. 

\subsubsection{Retrieval Training}

The goal of retrieval training is to develop an encoder that maps a given question $q$ and all relevant clues in the language space $\mathbb{Z}$ into an embedding space such that the question is close in proximity to its corresponding ground-truth evidence $z^+$. 
The objective is as follows:
\begin{equation}
        P_{retr}(z^+|q, \mathbb{Z})=\frac{{\rm exp}({\rm sim}(q,z^+))}{\sum_{z\in \mathbb{Z}} {\rm exp}({\rm sim}(q,z))}
\label{equ:retrieval2}
\end{equation}
where ${\rm sim}(q,z)$ is the cosine similarity between the normalized embeddings of the question and evidence, generated by the BERT encoder.
In order to perform contrastive learning, a set of negative evidences must be sampled as it is not feasible to enumerate all other evidences. 
This is done by using the BM25 algorithm to retrieve the most difficult negative clue for each positive clue and then placing them into batches of 32 instances. 
The training loss is then calculated as the negative log-likelihood for the positive evidence.

\subsubsection{Ranking Training}

The next phase of training, ranking, begins by gathering the initial retrieval results on the training set. 
The top 30 samples (excluding the ground-truth evidence $z ^+$) returned by the retrieval module are used as negative examples, and the ranker model is trained to distinguish positive cases from negative cases. 
In multimodal QA, where multiple positive evidences are required for reasoning, the negative of the summed log-likelihood for the positive evidences is used as the loss function. 
The logits $\mathbf{z}$ generated by the ranker are used as the clues and the indices for the correct evidences are taken from the ground-truth provenance $Pos$.
\begin{equation}
        \mathcal{L}_{ranking}=\sum_{i\in Pos}{\rm log}({\rm Softmax}(\mathbf{z})_i)
\label{equ:ranker2}
\end{equation}

\subsubsection{Generation Training}
Current generative models are trained to maximize the probability of generating the correct tokens at each decoding step. Given a question $q$, a set of collected clues $c$, and the ground-truth answer sequence $\mathbf{y}_{1:T}=\{y_1,\cdots,y_T\}$, the generation objective is to minimize the loss calculated by the following equation:
\begin{equation}
        \mathcal{L}_{generation}=\sum_{t=1}^T{\rm log}p(y_t|y_1,\cdots,t_t, q, c)
\label{equ:ranker2}
\end{equation}

\section{Experiments}

\subsection{Datasets}

\textbf{WebQA}~\cite{chang2022webqa} includes QA pairs that require one or two images and text snippets to answer. Each question has a set of distractors that the model must consider along with the correct clues to provide a answer. WebQA uses BARTScore to assess the fluency, and keyword accuracy to assess the correctness of the answer. These two scores are then multiplied to obtain the QA score.
The clue retrieval is easily evaluated via F1 score.


\textbf{MultimodalQA}~\cite{talmor2020multimodalqa} involves combining information from text, tables, and images to answer questions. 
Each question also includes visual and text distractors. 
The performance is measured by F1 score at the word level and the Exact Match (EM) of the predicted answer.


\textbf{MMCoQA}~\cite{li2022mmcoqa} is the first attempt at addressing the multi-modal conversational QA problem. 57.7\% of the conversations in the dataset involve two different modalities, while 24.4\% involve three modalities. 
The statistics all three datasets are demonstrated in Table~\ref{tab:statistics}.



\subsection{Implementation Details}

We conduct experiments on three datasets: WebQA, MMQA, and MMCoQA. The information source for WebQA includes both text and image modalities, while MMCoQA and MMQA focus on text, images, and tables.
Our method includes retrieval, ranking, and generation, though not all datasets require all three steps. 
For WebQA, a candidate clue list is given, and the model needs to find the most relevant clue to evaluate the accuracy of the clue retrieval, so a ranking module is necessary to sort the relevance.
Similarly, MMQA also provides a clue list, so retrieval is not necessary. 
Following previous work~\cite{li2022mmcoqa}, in MMcoQA, the positive clues are manually included in the retrieval results.
However, the number of candidate clues is not small (more than 20), which makes it difficult for the generation model to handle the long sequence, so a ranking model is needed to screen the candidate clues.
In all datasets,  the value of $K$ in retrieval is 30 and the value of $N$ in ranking is 10. 
The backbone for retrieval and ranking is BERT~\cite{kenton2019bert}, and the generation model is based on T5~\cite{raffel2020exploring}. 
We utilize the Transformers library and pre-trained parameters from HuggingFace~\footnote{\url{https://huggingface.co/}} and conduct experiments using 80G GPU cards. 
Further, AdamW~\cite{loshchilov2018decoupled} is used as the optimization algorithm with a a learning rate of 1e-4. 
The batch sizes for retrieval, ranking, and generation are 32, 3, and 1, respectively.
The image captioning is generated by BLIP~\cite{li2022blip}.
And the image-attribute features are obtained with VinVL~\cite{zhang2021vinvl}.

\subsection{Results}


\begin{table}
\centering
\footnotesize
    \resizebox{\linewidth}{!}{
        \begin{tabular}{lcccc}
        \toprule
                Model               & QA-FL       & QA-Acc     & QA  & Retr           \cr \midrule
                
                VLP       & 42.6 &36.7 &22.6     &68.9        \cr
                VLP + VinVL     & 44.2 &38.9 &24.1    &70.9   \cr
                MuRAG     & 55.7 &54.6 &36.1 &74.6   \cr
                \midrule
                Solar       & \textbf{60.9}   & \textbf{58.9}   & \textbf{40.9}    & \textbf{89.4}\cr
                \bottomrule 
        \end{tabular}
    }
\caption{WebQA official test-set results.}
\label{tab:webqa_result}
\vspace{-0.5em}

\end{table}

\begin{table}
\centering
\footnotesize
    \resizebox{\linewidth}{!}{
        \begin{tabular}{lcccc}
        \toprule
                Model               & QA-FL       & QA-Acc     & QA  & Retr           \cr \midrule
                VitaminC     & 59      & 57   & 39     & 84          \cr
                CMU ITL     & 60      & 58     & 39    &81     \cr
                HIT TMG   & 57   & 58   &39   & 89 \cr
                                \midrule
        Solar  & \textbf{61}   & \textbf{58}   &\textbf{41}   &\textbf{89} \cr
                
                \bottomrule
        \end{tabular}
    }

\caption{WebQA results indicated on leaderboard.}
\label{tab:webqa_result_leaber}
\vspace{-1.0em}
\end{table}

\textbf{WebQA.} We conduct experiments on the WebQA dataset, which includes both visual and textual data. Our results, presented in Table~\ref{tab:webqa_result}, demonstrate that Solar significantly outperforms the current SOTA method, MuRAG~\cite{chen2022murag}, across four metrics.
MuRAG encodes image patches and text tokens as sequences of vectors, which are then concatenated and fed to a decoder for answer generation. 
However, this approach requires pre-training the backbone with 300 million (image, text) pairs. 
In contrast, Solar converts images into text and performs QA in the language space, leveraging the world knowledge from public pre-trained models.
Comparing to the language pre-training corpus such as C4 (34 billion words)~\cite{raffel2020exploring}, multi-modal pre-training datasets are much smaller, which leads to less knowledge.

Furthermore, Solar ranks first on the WebQA leaderboard even without using model ensemble or additional post-processing strategies, as shown in Table~\ref{tab:webqa_result_leaber}. Note that the other methods we compared all employ external tricks to improve performance~\cite{yang2022enhancing}.

\begin{table}[t]
\centering { \footnotesize
\begin{tabular}{lcccccc} 
\toprule       
Model   & \multicolumn{2}{c}{Single-Modal} & \multicolumn{2}{c}{Mutli-Modal} & \multicolumn{2}{c}{All}
        \\ \cmidrule(r){2-3} \cmidrule(r){4-5} \cmidrule(r){6-7}
      & EM & F1 & EM & F1 & EM & F1   \\
\midrule                         
AR & 51.7 & 58.5 & 34.2 & 40.2 & 44.7 & 51.1 \cr 
ID & 51.6 & 58.4 & 44.6 & 51.2 & 48.8 & 55.5 \cr \midrule 
Solar & \textbf{69.7} & \textbf{74.8} & \textbf{55.5} & \textbf{65.4} & \textbf{59.8} & \textbf{66.1} \cr 
\bottomrule
\end{tabular}
\caption{MultimodalQA results.}
\label{tab:multimodalQA}
}
\vspace{-1em}
\end{table}

\textbf{MultimodalQA.} As demonstrated in Table~\ref{tab:multimodalQA}, Solar also achieves significant performance improvements on the more complex MultimodalQA dataset, which contain text, images, and tables. 
There are 11.0 pts and 20.6 pts improvements on EM and F1, respectively. 
The SOTA approach, ID~\cite{talmor2020multimodalqa}, requires an additional classifier to determine the appropriate modality for generating answers. 
This additional step can hinder the performance, as it relies on accurate classification results. 
In contrast, Solar blurs the boundaries between different modes at the input, allowing the model to generate answers from a unified text input without considering the modality type.


\begin{table}
\centering { 
\footnotesize
\begin{tabular}{lcccc} 
\toprule       
Model   & \multicolumn{2}{c}{Dev} & \multicolumn{2}{c}{Test} 
        \\ \cmidrule(r){2-3} \cmidrule(r){4-5}
      & EM & F1 & EM & F1   \\
\midrule                         
ORConvQA & 1.0 & 3.0  & 1.0 & 1.9 \cr
ManyModelQA & 0.7 & 2.3 & 1.0 & 1.8 \cr 
MAE & 21.5 & 30.2 & 24.9 & 32.3 \cr \midrule 
Solar & \textbf{56.8} & \textbf{62.5} & \textbf{57.3} & \textbf{64.6} \cr
\bottomrule
\end{tabular}
\caption{MMCoQA test-dev-set results.
}
\label{tab:mmcoqa}
}
\vspace{-1.0em}
\end{table}

\textbf{MMCoQA.} Solar demonstrates the highest performance improvement on the most challenging MMCoQA dataset (Table~\ref{tab:mmcoqa}). 
The EM and F1 scores have both increased by 32.4 and 32.3 pts, respectively. 
In comparison to MultimodalQA, MMCoQA requires the model to correctly incorporate dialog history. 
The previous SOTA method MAE~\cite{li2022mmcoqa}, relies on ensembling three conventional QA models and utilizing a modality classifier to determine the final output. 
However, MAE lacks the ability to reason between modalities, which is crucial for MMCoQA, as 35.7\% of the questions require cross-modal reasoning. 
Our proposed Solar is able to effectively retrieve multiple relevant clues in the unified language space (potentially belonging to different modalities), and utilize cross attention in transformer decoding to perform cross-modal reasoning.

In addition, Solar also significantly reduces the feature storage. Text storage is more efficient than image feature, For example, AR requires 78G of space to store Faster R-CNN~\cite{ren2015faster} extracted image features, while Solar only requires 186M of space to store text representation of three modalities, thus saving 99.7\% of storage space. This is particularly important for online services, as feature extraction, such as converting images to embeddings or text, can be done offline, but feature storage is always present.We also give an ablation study in \textbf{Appendix}~\ref{sec:ablation}.
Our analysis shows that both global and local textualization of images have a positive effect on the results, with global features being more significant.



\section{Conclusion}

In this paper, we approach the multi-modal QA task from a new perspective, where all the information sources are aligned into the language space to take advantage of the power of pre-trained language models and facilitate cross-modal reasoning. 
Our experiments show the promise of this approach, as it outperforms baselines by a large margin and achieves the top position at public leaderboard with a single model.
We hope this work can lead to a discussion: can language become a unified interface for people and models to understand the world?

\section{Limitations}

While our Solar has demonstrated its superior performance on three benchmarks, it still has several limitations.
Firstly, Solar relies on accurate recognition of image caption and object-attribute detection models. 
If the features of these two parts are not correctly recognized, it will cause subsequent cascading errors. 
Secondly, it only demonstrates that language can serve as a unified representation in multi-modal QA, but has not been tested in other more multi-modal tasks, which we will leave as future work.
Lastly, the experimental results do not delve deeper into which cases a unified language representation is better and in which cases a multi-modal model performs better. We speculate that an integration of language models and multi-modal models will yield better results.

\section{Ethics Statement}

Our submission adheres to the ACL Code of Ethics as it utilizes open public datasets and models, and does not raise any ethical concerns regarding human subjects, harmful insights, conflicts of interest, discrimination, privacy, security, legal compliance, or research integrity. 
Additionally, we have thoroughly referenced related works and have compared our results accordingly.

\bibliography{anthology,official_custom}
\bibliographystyle{acl_natbib}

\appendix

\label{sec:appendix}

\section{Related Work}
\label{sec:related}

\subsection{Multimodal Question Answering} 
The problem of multimodal question answering has been extensively studied. 
VQA~\cite{antol2015vqa} is firstly proposed to answer questions from visual-only inputs. 
Later, WebQA~\cite{chang2022webqa}, MuMuQA~\cite{reddy2022mumuqa}, ManyModalQA~\cite{hannan2020manymodalqa} provide questions which require reasoning over images and explicitly provided text snippets. 
More recently, MultimodalQA~\cite{talmor2020multimodalqa} and MMCoQA~\cite{li2022mmcoqa} require integrating information across free text, semi-structured tables, and images.
To address the challenge of finding answers from multiple sources of information, MuRAG~\cite{chen2022murag} designs a multi-modal transformer architecture to accept both text and image feature inputs, and builds a million-scale dataset for pre-training the model. 
AutoRouting~\cite{talmor2020multimodalqa} and MAE~\cite{li2022mmcoqa} train separate models for each modality, fully utilizing existing pre-trained models and then using classifiers to determine which modality to answer a question from, but unable to perform cross-modal reasoning. 

In contrast to these two methods, which perform multi-modal QA within the multi-modal space, we propose Solar, which transforms tables and images to the language space and uses a unified pre-trained language model to achieve multi-modal QA. 
This brings two advantages: first, it fully utilizes the language model pre-trained on tens of billions of tokens, having more world knowledge than multi-modal models with only a few million pre-training data; 
second, it can naturally achieve cross-modal reasoning. 
Experiments on three datasets have demonstrated the advantages of Solar in terms of performance, parameter amount, and storage space.

\subsection{Unified Model} 
One of the most ambitious goals in the field of artificial intelligence is the development of a unified model that can perform a wide range of tasks across different modalities~\cite{lu2022unified,ning2023all,hao2022language,cao2023towards}. 
However, this is a challenging task due to the diversity and complexity of real-world scenarios. 
Despite these difficulties, recent advances in natural language processing, specifically with large-scale language models such as GPT-3, have demonstrated impressive capabilities as a general-purpose solution for language-based tasks. 
This success has encouraged researchers to investigate the potential for universal models in other areas, such as computer vision, through works like knowledge-based VQA~\cite{yang2022empirical,changpinyo2022all,gao2022transform}.
In this paper, we explore an even more complex multi-modal task: finding answers from three different data sources, including text, images, and tables. We demonstrate that language can be used as a unified interface to solve such a complex task.

\subsection{Retrieval-augmented Generation} 

Solar is also closely related to retrieval-augmented generation. 
This approach combines the benefits of retrieval and generation methods to provide more comprehensive and accurate responses to complex questions~\cite{lewis2020retrieval,mao2021generation}. Retrieval-augmented generation models have been shown to outperform standalone language models by leveraging a knowledge base for additional context and provenance~\cite{hofstatter2022fid, glass2022re2g,zhao2023causal}. 
The results of Solar demonstrate the effectiveness of this approach in multi-modal QA.

\begin{table}[t]\small
	\centering
	\resizebox{\linewidth}{!}{
	\begin{tabular}{lcccccc}
		\toprule
		  Datasets & Text? & Table?  & Image? & Retrieve? & Rank?  & Generate?  \\
		\midrule
		WebQA  & \ding{51} &\ding{55} &\ding{51} &\ding{55}  & \ding{51} & \ding{51} \\
		MMQA & \ding{51} &\ding{51} &\ding{51} &\ding{55}  & \ding{51} & \ding{51}\\
		MMCoQA & \ding{51} &\ding{51} &\ding{51} &\ding{51}  & \ding{55} & \ding{51}\\
		\bottomrule
	\end{tabular}}
		\caption{Comparison of Multi-modal QA datasets.  }
	\label{tab.comp}
\end{table}

\begin{table}[t]
	\centering
\small

	\begin{tabular}{lccc}
		\toprule
		  Datasets & \#Train & \#Dev  & \#Test   \\
		\midrule
		WebQA  & 34.2k & 5k & 7.5k  \\
		MultiModalQA & 23.8K & 2.4K & -  \\
		MMCoQA & 45.8k & 0.6k & 0.6k\\
		\bottomrule
	\end{tabular}
		\caption{Overall Statistics of downstream datasets.  }
	\label{tab:statistics}
 \vspace{-1em}

\end{table}

\section{Ablation Study}
\label{sec:ablation}

\begin{table}[htb]
\begin{center}
\small
\begin{tabular}{lcc}
  \toprule
      Model                 & EM & F1  \\
  \midrule
  Solar                & 56.8 & 62.5     \\
  \quad-- Global textualization  & 48.9 & 54.1   \\
  \quad-- Local textualization  & 56.3 & 60.9     \\
  \bottomrule
\end{tabular}
{\caption{Ablation study of Solar on MMCoQA dev set.}
\label{tab:ablation}}
\end{center}
\vspace{-1em}
\end{table}


In this experiment, we ablate the Global and Local textualization strategies for images. 
Note that we cannot ablate retrieval, ranking, and generation as all datasets require generation, and each dataset only contains one of the steps of retrieval or ranking. 
Linearization of tables also cannot be ablated otherwise tables cannot be mapped to the language space. 
Results of the ablation experiments are shown in Table~\ref{tab:ablation}.
It can be seen that both global and local features have a positive impact on the results, with global features being particularly significant. 
We hypothesize that this may be partly due to the importance of image captions for understanding the overall semantic meaning of an image, and partly due to the fact that image titles are often relevant answers to many visual-related questions.

\end{document}